\documentclass{article}
\usepackage{multirow}

\usepackage{PRIMEarxiv}
\usepackage{subfigure}
\usepackage[utf8]{inputenc} 
\usepackage[T1]{fontenc}    
\usepackage{hyperref}       
\usepackage{url}            
\usepackage{booktabs}       
\usepackage{amsfonts}       
\usepackage{nicefrac}       
\usepackage{microtype}      
\usepackage{lipsum}
\usepackage{fancyhdr}       
\usepackage{graphicx}       
\graphicspath{{media/}}     
\title{TechGPT-2.0: A large language model project to solve the task of knowledge graph construction
} 

\author{
  Wang Jiaqi*, Chang Yuying*, Li Zhong*, An Ning*, Ma Qi*, Hei Lei*, Luo Haibo, Lu Yifei, Ren Feiliang \\
  Northeastern University \\
  ShenYang, China\\
  \texttt{wangjiaqi@stumail.neu.edu.cn, 2201766@stu.neu.edu.cn} \\
  \texttt{2272001@stu.neu.edu.cn, luohaibo@stumail.neu.edu.cn} \\
  \texttt{2301922@stu.neu.edu.cn, renfeiliang@cse.neu.edu.cn,} \\
}

\begin{document}
\maketitle

\begin{abstract}
Large language models have exhibited robust performance across diverse natural language processing tasks. This report introduces TechGPT-2.0, a project designed to enhance the capabilities of large language models specifically in knowledge graph construction tasks, including named entity recognition (NER) and relationship triple extraction (RTE) tasks in NLP applications. Additionally, it serves as a LLM accessible for research within the Chinese open-source model community. We offer two 7B large language model weights and a QLoRA weight specialized for processing lengthy texts.Notably, TechGPT-2.0 is trained on Huawei's Ascend server. Inheriting all functionalities from TechGPT-1.0, it exhibits robust text processing capabilities, particularly in the domains of medicine and law. Furthermore, we introduce new capabilities to the model, enabling it to process texts in various domains such as geographical areas, transportation, organizations, literary works, biology, natural sciences, astronomical objects, and architecture. These enhancements also fortified the model's adeptness in handling hallucinations, unanswerable queries, and lengthy texts. This report provides a comprehensive and detailed introduction to the full fine-tuning process on Huawei's Ascend servers, encompassing experiences in Ascend server debugging, instruction fine-tuning data processing, and model training. Our code is available at \url{https://github.com/neukg/TechGPT-2.0}
\footnote{* The authors contribute equally, all are co-first authors of the article and the names are randomly ordered.}
\footnote{Our code is available at \url{https://github.com/neukg/TechGPT-2.0}}
\footnote{Model weights are available at \url{https://huggingface.co/neukg/TechGPT-2.0-alpaca-hf} and \url{https://huggingface.co/neukg/TechGPT-2.0-atom-hf}}
\end{abstract}


\section{Introduction}
The current discussion on the relationship between large language models (LLMs) and knowledge graphs (KGs) is highly active. Whether these two can complement each other, how LLMs and KGs can be effectively integrated, and the question of whether future research should focus on KG assisting LLMs or vice versa remain unanswered. To advance research on the amalgamation of knowledge graphs and large models, we initiated the TechGPT-1.0 and TechGPT-2.0 projects. These projects encompass the release of three 7B-scale instruction fine-tuning models and a QLoRA model tailored for long texts. The extensive models developed within the TechGPT-1.0 and TechGPT-2.0 projects concentrate primarily on diverse subtasks related to constructing knowledge graphs, including tasks like NER and RTE. Furthermore, we will furnish comprehensive insights into data collection and processing, share experiences in debugging the Ascend server, and delineate our model training procedures. This information aims to serve as a valuable reference and support for researchers seeking to train large-scale language models on the Ascend server.

Specifically, experienced researchers in our research group curated batches of knowledge graphs to construct datasets for subtasks such as NER and RTE. These datasets were subsequently modified to align with instructions and fine-tune the data format. Notably, the knowledge graph employed for constructing the subtask dataset encompasses open-source data collected from public datasets, alongside research data accumulated in previous projects of our research group. This data undergoes collation and manual annotation by our group's researchers.

To ensure the sustained general ability of the large language model during the complete fine-tuning process, we adhere to conclusions drawn from prior experiments. A substantial volume of general instruction fine-tuning data is compiled, and this is amalgamated with knowledge graph domain data in appropriate proportions, resulting in an approximate 4 million instances of instruction fine-tuning data. Additionally, significant disparities exist between the model training process on the Ascend server and traditional NVIDIA graphics cards. Our intention is to share insights into the unique challenges and practices related to model training on the Ascend server, offering valuable guidance for researchers contemplating exploration on this platform in the future.

Owing to limitations in our server resources, detailed experiments on the models within the TechGPT-1.0 and TechGPT-2.0 projects were not pursued. Consequently, this report provides only concise summaries of straightforward experimental outcomes. For a more in-depth exploration of model performance, we encourage you to visit open-source communities like HuggingFace, ModelScope, and WiseModel for downloads and firsthand experiences. Alternatively, you may explore the detailed experiences available on the experience page shared by our research team.

\section{Related Work}
\subsection{Large Language Model}
With the rapid development in the field of natural language processing (NLP), Large Language Models (LLMs) have garnered widespread attention in recent years. Language modeling, as a primary approach for processing natural language text, has been extensively researched over the past two decades, utilized for language understanding and generation. It has evolved from traditional statistical language models to neural language models and, more recently, has gained substantial focus on pre-training language models. The concept of pre-training originated from transfer learning \cite{yosinski2014transferable} in Computer Vision (CV) tasks. When applied to NLP, pre-training techniques involve training Transformer \cite{vaswani2017attention} models on large-scale corpora, enabling the capture of rich knowledge beneficial for downstream tasks, such as long-term dependencies and hierarchical relationships. In addition, the significant advantage of pre-training in the field of NLP is that the training data can be derived from any unlabeled text corpus, providing virtually limitless training data for the pre-training process. Early pre-training methods were static, such as Neural Network Language Models (NNLM) \cite{bengio2000neural} and Word2Vec \cite{mikolov2013efficient}. However, static methods struggled to adapt to diverse semantic contexts. Consequently, dynamic pre-training techniques, such as BERT \cite{devlin2018bert} and XLNet \cite{yang2019xlnet}, were introduced. As hardware capabilities rapidly advanced, researchers discovered that increasing parameters and training data scale of pre-training language models could lead to substantial performance improvements.

The GPT series, representing a paradigm in LLMs, have undergone multiple versions of iterations and shown significant development trajectories. From the initial attempts with GPT-1 \cite{radford2018improving} and GPT-2 \cite{radford2019language} to the substantial leap in model capabilities achieved by GPT-3 \cite{brown2020language}, and further advancements seen in GPT-4 \cite{openai2023gpt4}, which supports multimodal input and possesses enhanced comprehensive abilities. Recently, the evolution of these series has not only demonstrated notable performance improvements in natural language processing tasks but has also contributed significantly to the research on LLMs. The emergence of ChatGPT, in particular, further substantiates the allure of large-scale language models. These large-scale pre-training language models are called large language models, such as LLaMA \cite{touvron2023llama} (65B), GPT-3 (175B), BLOOM \cite{workshop2022bloom} (176B), and PaLM \cite{chowdhery2023palm} (540B).

Since the beginning of 2023, various large language models from both the industry and research institutions one after another. It is worth mentioning that Chinese large language models have also sprung up like mushrooms after rain in the past one year. Among them, ChatGLM \cite{du2021glm,zeng2022glm} is one of the most effective open-source base models in the Chinese field, which has been optimized for Chinese question answering and dialogue. Within a mere year, three versions of the model have been trained. On the basis of retaining many excellent features such as smooth dialogue and low deployment threshold of the previous two generations of models, ChatGLM3 also extends the context length of the base model to 32K, and trains with 8K context length during the dialogue phase, allowing free commercial use. Additionally, Alibaba Cloud has developed the Qwen \cite{bai2023qwen} series of general-purpose question-answering large models, encompassing parameter scales of 1.8 billion (1.8B), 7 billion (7B), 14 billion (14B), and 72 billion (72B). These models support an 8K context length and feature specific optimizations for aligning data related to plugin calls. The current models demonstrate effective plugin invocation and are upgradeable to an Agent.

The rapid development of LLMs has sparked significant interest in harnessing their potential across various natural and social science domains to address domain-specific tasks. However, some challenges, such as limited domain-specific expertise, knowledge induction, and model complexity, represent pressing obstacles to the direct application of LLMs in these domains.

\subsection{Instruction Tuning Paradigm}
The training of large language models typically involves two stages: (1) unsupervised pre-training on massive unlabeled text corpora, learning universal semantic representations and world knowledge, and predicting the next token in the token sequence; (2) fine-tuning on small-scale data, incorporating prompt engineering and Reinforcement Learning based on Human Feedback \cite{stiennon2020learning} (RLHF) to better align the final task with human preferences. LIMA \cite{zhou2023lima} has demonstrated that almost all of LLM's knowledge is learned during pre-training, and high-quality responses can be generated with limited instruction fine-tuning of data. Consequently, the performance of the base model is crucial, as effective fine-tuning and reinforcement learning heavily depend on the base model's proficiency. Since instruction fine-tuning does not inject new capabilities into the model, but rather unlocks or activates existing ones, the foundational abilities are already established during the pre-training phase. This is primarily due to the significantly smaller amount of data available for fine-tuning compared to the pre-training phase. For instance, the knowledge in the training corpus is mostly declarative sentences, such as "The capital of China is Beijing". If the prompt "The capital of China is" is given to the large model during pre-training, it can effortlessly complete the answer with "Beijing". However, if the question takes the form of an interrogative sentence like "Which city is the capital of China?", even though the large model likely can answer this simple question with just pre-training, if the question becomes more complex, the large model may struggle to provide a satisfactory response. In such cases, even though the answer may be present in the pre-training corpus, fine-tuning is necessary to fully exploit the potential of the large model.

From the perspective of parameter scale, the instruction fine-tuning of LLMs primarily falls into two technical pathways: (1) Full Fine Tuning (FFT), which involves training all parameters with the entire dataset; (2) Parameter-Efficient Fine Tuning \cite{houlsby2019parameter} (PEFT), which entails training only a subset of parameters. The principle behind Full Fine Tuning is to train all parameters of the LLM using specific data, offering the primary advantage of superior performance in a specific domain. However, Full Fine Tuning often faces two significant issues: (1) the training cost is relatively high due to the volume of parameters being equivalent to pre-training; (2) Catastrophic forgetting \cite{mccloskey1989catastrophic}, where fine-tuning on specific data may enhance performance in that domain but potentially degrade capabilities in previously well-performing domains. PEFT aims to address the aforementioned issues associated with FFT and is currently a prevalent fine-tuning approach. Some popular EFT strategies include: (1) Prompt Tuning \cite{lester2021power}, which maintains the base model's parameters and trains a small model with few parameters for each specific task, invoked as needed during task execution. (2) Prefix Tuning \cite{li2021prefix}, based on the observation from Prompt Engineering that adding appropriate conditions in the prompt context without altering the large model can guide it to achieve superior performance. While the motivation behind Prefix Tuning is similar to Prompt Tuning, their specific implementations differ. Prompt Tuning adds specific tokens to the input sequence during the embedding stage, whereas Prefix Tuning introduces specific prefixes in both the Encoder and Decoder networks of the Transformer. (3) LoRA \cite{hu2021lora}, takes a completely different technical pathway from Prompt Tuning and Prefix Tuning. LoRA's premise is rooted in the assumption that current large language models are excessively parameterized. Behind overparameterized large models lies an essential low-dimensional model, suggesting that while large models have many parameters, not all parameters play equal roles. Some parameters in large models are crucial and significantly impact the generated results; these crucial parameters represent the low-dimensional essence mentioned earlier. (4) QLoRA \cite{dettmers2023qlora}: While LoRA has demonstrated remarkable effectiveness, almost comparable to Full Fine Tuning, the introduction of QLoRA is primarily driven by the quantization aspect. Quantization is a method that reduces the precision of parameters, thereby lowering the computational resource requirements of the model, while ensuring minimal impact on model performance. QLoRA is a quantized version of LoRA, further reducing the bit representation from 16 bits to 4 bits, aiming to significantly reduce costs while preserving model effectiveness.

\label{sec:headings}




\section{Methodlogy}
\label{sec:others}
This section presents various aspects of the TechGPT project, encompassing model configurations, intricacies of data collection and processing, debugging experiences with the Ascend server, and insights gained from model training. The subsection on model settings delves into challenges encountered during model selection in the preliminary research phase. Regarding data collection and processing, we elaborate on the composition of the 4 million instruction fine-tuning dataset, intricacies of data processing, and the data construction process. The section discussing the debugging of the Shengteng server details its usage and highlights encountered challenges. In the realm of model training experiences, we address issues such as the influence of data on training and propose solutions to challenges associated with processing long texts. The main hyperparameters can be found in Table \ref{tab:table1}.





\subsection{Model settings}
The TechGPT project encompasses two distinct initiatives, TechGPT-1.0 and TechGPT-2.0. TechGPT-1.0 features a 7B scale model, while TechGPT-2.0 incorporates two 7B scale models and a QLoRA model weight. The model in TechGPT-1.0 employs a Chat model from BELLE as the original framework, and multiple models were explored during the development phase of the TechGPT-2.0 project. Training transpired on the Ascend server, utilizing the Mindspore and Mindformer libraries for debugging. Both toolkits and the Ascend server firmware exclusively support the training of models using LLAMA\cite{touvron2023llama} and ChatGLM \cite{zeng2023glm130b} as the foundational architecture. Consequently, an investigation ensued, exploring models like LLAMA, LLAMA2\cite{touvron2023llama}, ChatGLM, ChatGLM2, PanGuAlpha\cite{zeng2021pangualpha}, Baichuan, Baichuan2\cite{yang2023baichuan}, Qwen, Internlm \cite{team2023internlm}, and others supported by Mindformer. Among these, the LLAMA-like model and the ChatGLM-like model demonstrated superior performance. PanGuAlpha, Baichuan, and Baichuan2 encountered debugging challenges due to equipment and code issues. The remaining models may exhibit suboptimal performance compared to the LLAMA-like model and the ChatGLM-like model post fine-tuning, attributed to adaptation issues. Simultaneously, during the fine-tuning process of the LLAMA-like model, potential adaptation challenges surfaced among the Mindformer version, Mindspore version, and server firmware version, resulting in a comparatively poor outcome post fine-tuning for the LLAMA2 base model. Only the Chat version model underwent complete fine-tuning, showcasing commendable performance. Following the full fine-tuning of the ChatGLM model on specific data, its performance excelled solely on the knowledge graph construction subtask data within the training set, yet it failed to demonstrate a discernible level of generalization ability. Concurrently, the ChatGLM model proved incapable of showcasing knowledge graph construction ability following LoRA fine-tuning on all data. Furthermore, the ChatGLM model exhibited a lack of generalization ability after LoRA fine-tuning on the knowledge graph construction task data, resulting in a complete loss of all general capabilities.

\begin{table}
 \caption{Hyperparameter Settings}
  \centering
  \begin{tabular}{lll}
    \toprule
    \multirow{2}{*}{hyperparameter} & \multicolumn{2}{c}{Model} \\
    \cmidrule(r){2-3}
       & Atom   & Alpaca    \\
    \midrule
    epochs & 1  & 1 \\
    batch\_size     & 2  & 2\\
    learning\_rate     & 8e-6     & 1.5e-7   \\
    warmup\_ratio     & 0.03    & 0.01     \\
    seq\_length     & 4096     & 4096     \\
    hidden\_size     & 4096    & 4096      \\
    num\_layers     & 32     & 32    \\
    num\_heads      & 32     & 32    \\
    vocab\_size    & 65000    & 55296     \\
    pretrain\_seqlen     & 4096      & 4096    \\
    \bottomrule
  \end{tabular}
  \label{tab:table1}
\end{table}

In summary, following analysis, it was determined that the Ascend server is well-suited for the LLAMA and LLAMA2 models. Consequently, the decision was made to designate LLAMA2 as the original model for the TechGPT-2.0 project. With the aim of providing a substantial model for the Chinese Internet within the knowledge graphs domain, focus was narrowed to the LLAMA2 model from the Chinese open-source community. Further investigation delved into the Chinese LLAMA2 model released by HFL and the Atom model jointly released by the Llama Chinese community and AtomEcho. Subsequent to less-than-optimal model results following full fine-tuning using the base model in early experiments, attention turned to experiments involving the Chat versions of the aforementioned models. Within the Chinese LLAMA2 model by HFL, assessments were conducted on the Chat version of the Chinese-Alpaca-2 (7B) model and the Chinese-Alpaca-2-16K (7B) model, particularly focusing on long text. Anticipating capabilities related to long text processing in the TechGPT-2.0 version, numerous experiments were conducted on the Chinese-Alpaca-2-16K (7B) model; however, the outcomes were unsatisfactory. The Chinese-Alpaca-2-16K (7B) model exhibited poor results in direct inference output. Efforts were directed toward optimizing its output results through subsequent instruction fine-tuning. Unfortunately, during subsequent fine-tuning, the loss consistently increased sharply, preventing model convergence. The checkpoint before convergence was utilized for reasoning, showcasing specific abilities only in the NER task, but lacking proficiency in answering RTE task questions.

Ultimately, the decision was made to employ Chinese-Alpaca-2 (7B) and Atom-7B-Chat as the initial models for the TechGPT-2.0 project. The model underwent fine-tuning with QLoRA\cite{dettmers2023qlora} using the position interpolation method to enhance its capacity for long-text processing. Both HFL and AtomEcho expanded and optimized the Chinese vocabulary of the original LLAMA2. Large-scale Chinese data were utilized for incremental pre-training, resulting in their respective Chinese LLAMA2 base models. Subsequent fine-tuning instructions were applied based on these Chinese LLAMA2 base models to derive our selected Chat versions. This technical approach not only enhances the fundamental Chinese semantics and instruction understanding capabilities of the large model but also improves its efficiency in processing Chinese text. Simultaneously, this approach preserves the original LLAMA2 model's ability to process 4k text, facilitating further research and experimentation on long-text problems. It is noteworthy that, despite both Chat models employing the same technical solution, their performance differs. This disparity may be attributed to the superior pre-training data and instruction fine-tuning data used by Chinese-Alpaca-2 (7B) compared to Atom-7B-Chat. Chinese-Alpaca-2 (7B) exhibits slightly better overall capabilities, particularly in tasks such as summary questions and dialogue. However, Atom-7B-Chat demonstrates a potential advantage in answering a broader range of questions due to the superior diversity of its pre-training and instruction fine-tuning data. Importantly, the distribution of our data aligns closely with the instruction fine-tuning data distribution used by Atom-7B-Chat. During instruction fine-tuning on Chinese-Alpaca-2 (7B), sporadic Loss surges were observed, preventing model convergence. In the initial capability test on both models, an additional test focused on the knowledge graph construction subtask. Chinese-Alpaca-2 (7B) exhibited partial capabilities in the NER task, while Atom-7B-Chat performed poorly. Both models demonstrated suboptimal performance in the RTE task. This outcome is encouraging as it substantiates that current small-scale large models lack the ability to construct knowledge graphs. Our research aims to address this gap in the Chinese open-source community's exploration of this field.



\subsection{Data collection and processing}
\begin{figure}[ht]
    \centering
    \includegraphics[scale=0.3]{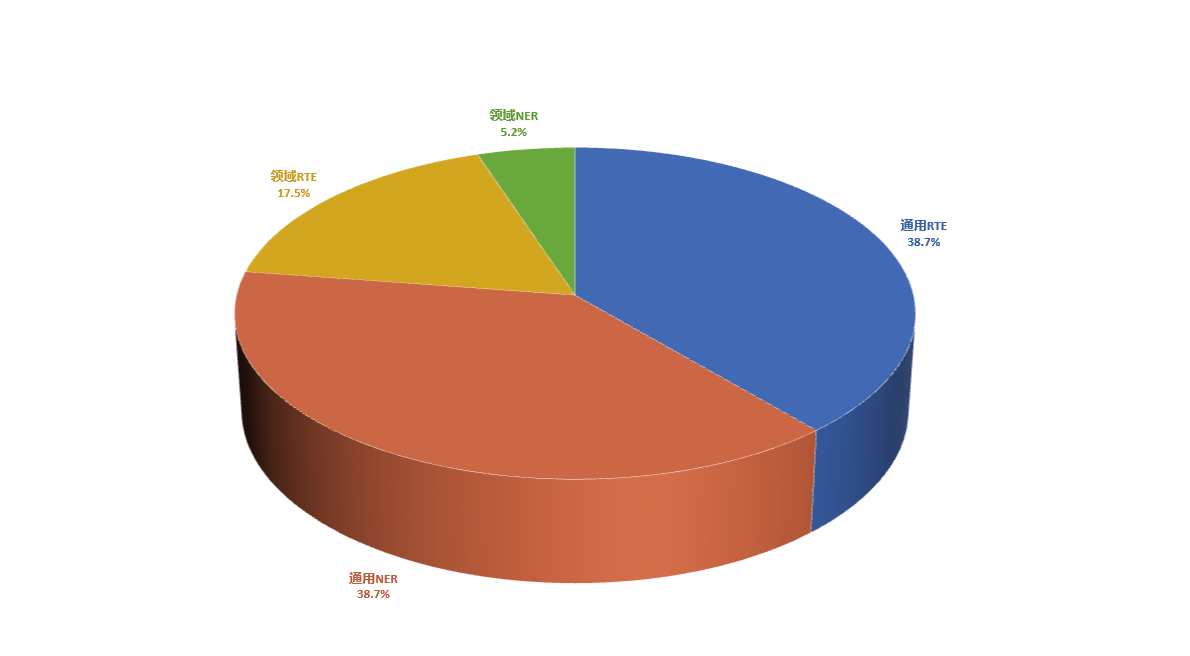}
    \caption{RTE Data Proportion Chart}
    \label{fig:fig1}
\end{figure}

The collection of our dataset is delineated into two stages, aligning with the two projects of TechGPT. The initial stage encompasses the first set of data, where approximately 3.8 million instances of instruction fine-tuning data were gathered and organized. Subsequently, in the second stage, an additional 220,000 instances of instruction fine-tuning data were amassed, resulting in a cumulative dataset of approximately 4 million instances. In the subsequent section, a detailed exposition of the distribution of these 4 million instances will be presented. The proportion of RTE data is shown in Figure \ref{fig:fig1}.

The focus of the TechGPT project lies in the knowledge graph construction task, comprising two primary subtasks: named entity recognition and entity relationship triplet extraction. The 400W instruction fine-tuning dataset encompasses 15W instances of general RTE task data, 15W instances of general NER data, and 6.8W instances of domain RTE task data. This includes 1.8W instances of medical RET data and 20W instances of domain NER data. In summary, the knowledge graph construction task dataset totals 38.8W instances. The general RTE task data and general NER data originate from the early research accumulation and annotated collections of our research group, subsequently transformed into instruction fine-tuning data through straightforward rules. Most of these early accumulated data were manually and rule annotated by members of our research team. The data accumulated in the early stages underwent a certain level of screening, resulting in the final selection of 30W instances for instruction fine-tuning. Domain RTE task data and domain NER task data partly originate from open-source instruction fine-tuning datasets within the Chinese community, such as the KnowLM-IE \cite{zhang2024comprehensive} dataset provided in the KnowLM open-source project. Additionally, part of the data comes from annotated datasets collected during research group cooperation projects. The NER data in these domains includes some nested NER data. It is noteworthy that the KnowLM-IE dataset is annotated through remote supervision, introducing considerable noise. Directly amalgamating it with our data would compromise the model's ability to extract entity relationship triples.

In addition to the data incorporated into the knowledge graph construction task, efforts were made to preserve the model's proficiency across various tasks by amassing an extensive corpus of general task and dialogue data. This compilation encompasses approximately 3 million instances of general dialogue data and spans over ten diverse tasks, including 3.5W Chinese-English translations, 3.5W English-Chinese translations, constituting a total of 70K instances of Chinese-English translation task data. Other task data consists of 15W title expansion summaries, 20W abstract abbreviation titles, 1W code instruction fine-tuning instances, 10W keyword extraction instances, and 3W subject area identification instances, among others. To align the model's output more closely with human values, additional datasets were collected, encompassing value alignment data, hallucinations, and unanswerable question data. This comprises 1,000 instances of unanswerable question data and 0.3,000 instances of value alignment data. Notably, the unanswerable question data includes both positive and negative responses. Furthermore, QA data was gathered from domains such as medicine and law, encompassing 100,000 instances of medical QA data and approximately 30,000 instances of legal QA data. The legal QA data encompasses legal case sorting and legal consultation QA data, while the medical QA data primarily comprises medical consultation QA instances.

\begin{figure}[ht]
    \centering
    \includegraphics[scale=0.3]{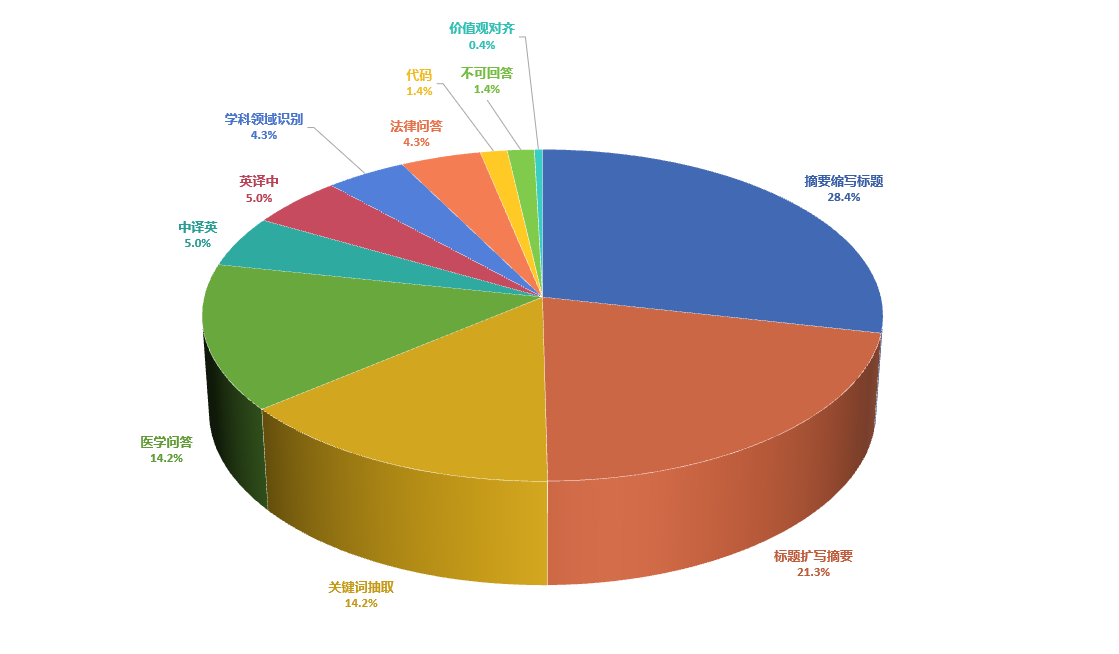}
    \caption{General task data proportion chart}
    \label{fig:galaxy}
\end{figure}

Finally, we compiled approximately 4 million instructions of fine-tuning data for fine-tuning the two 7B models selected in the TechGPT-2.0 project.

\begin{figure}[ht]
    \centering
    \includegraphics[scale=0.3]{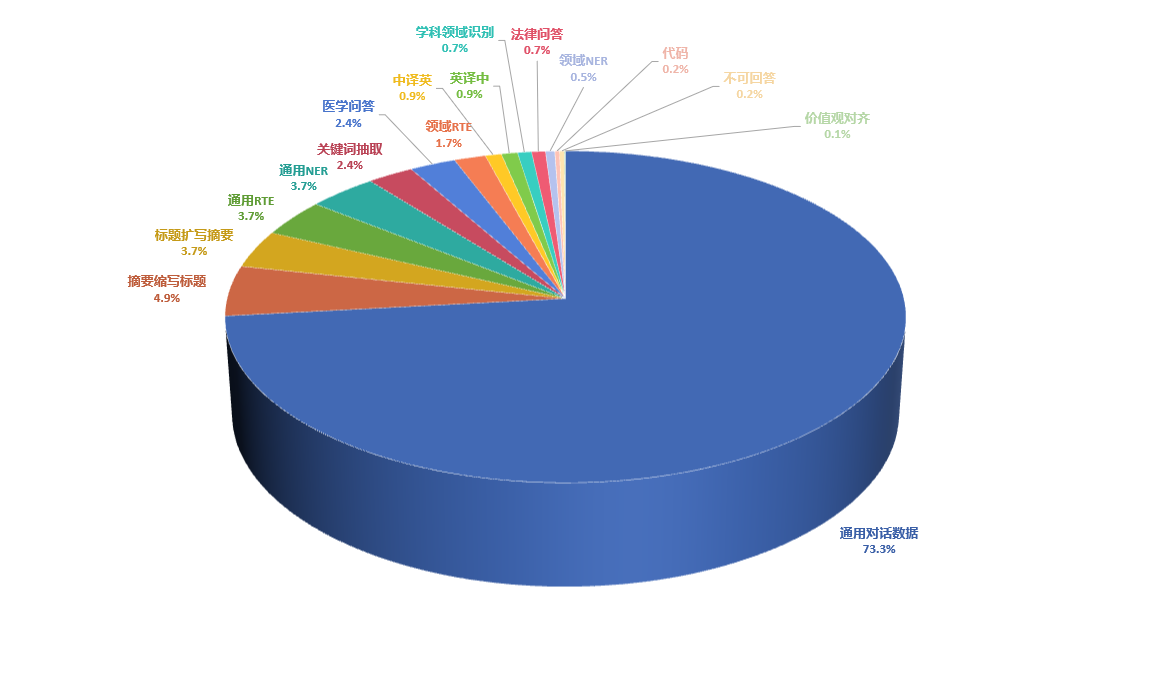}
    \caption{Summary data proportion chart}
    \label{fig:galaxy}
\end{figure}



\subsection{Ascend server debugging}
In the TechGPT-2.0 project, owing to the non-local deployment of the Ascend server, communication with five Ascend 910A servers was facilitated through a springboard machine. Huawei provides two official debugging connection methods: one utilizing the bare metal approach for communication with five Ascend 910A servers, and the other leveraging the Modelarts tool to establish communication. Given the necessity for detailed model operations, we opted for the former method, the bare metal solution. It is noteworthy that we possess 5 Ascend 910A servers, with only 4 allocated for actual training purposes. This decision stems from potential issues related to communication, data, and model parallel computing observed during the training process. When utilizing the same model, identical data volume, and matching configurations, the training efficiency with 5 servers was found to be lower than that achieved with 4 servers.

The TechGPT-2.0 project employs the Huawei Ascend 910A server, Mindspore computing framework, and Mindformer suite for its development. A distributed computing solution is implemented, utilizing a total of 4 machines with 8 NPUs each (32G) for training. The specific versions used include Ascend server firmware CANN version 6.3 rc2, Mindspore-2.1.1, and Mindformer-dev. It is noteworthy that during the development of the TechGPT-2.0 project, the Mindspore computing framework and Mindformer suite versions underwent rapid updates and iterations, resulting in potential version control challenges. During this period, LLAMA2 had recently been adapted on the Ascend server, introducing subtleties that could impact model training. The aforementioned versions represent a relatively stable set that successfully facilitated full fine-tuning of the LLAMA2 model after thorough testing. Subsequently, we delve into detailed experiences in debugging on Ascend servers.

\paragraph{Data issues} The utilization of the Chat version of the large model as the initial model necessitates the assurance of consistency in our instruction fine-tuning data with the structure used during Chat model training. Chinese-Alpaca-2 (7B) adheres to the same Chinese Alpaca command format as LLAMA2-Alpaca, while Atom-7B-Chat adopts a command format akin to that of LLAMA1-Alpaca. When employing Mindformer for training, data conversion into the Mindrecord format is imperative. Throughout this conversion process, additional instructions are incorporated into the dataset, mirroring the instruction format of LLAMA1-Alpaca. Subsequent testing revealed optimal model performance when removing the extra instructions introduced by Mindformer, ensuring alignment with the instruction fine-tuning data structure used in training the corresponding Chat model. The inclusion of these supplementary instructions or the use of inconsistent instruction formats heightens the likelihood of the model's failure to converge to some extent. Even in cases where convergence is achieved, substantial variations in reasoning performance may arise across different tasks when employing distinct instruction formats.

\paragraph{Model and parameter setting issues} 

Upon migration of Mindformer's LLAMA2 project from LLAMA1, a certain degree of inadequate adaptation is observed, making it challenging at times to ascertain whether issues arise from adaptation challenges, the model, or the data itself. Here, we share select experiences encountered during the training process. In the initial phase of project development, Mindformer lacked adaptation to LLAMA2, necessitating modifications based on the configuration file of LLAMA1. Key parameters, such as learning rate, warmup ratio, sequence length, and vocabulary size, must be adjusted to align with the selected configuration file for the LLAMA2 initial model. However, due to adaptation challenges, instances of insufficient video memory may persist. Specifically, vocabulary size must be configured differently based on the selected model, requiring corresponding changes to the special token identifier. The learning rate should be minimized during full fine-tuning, with specific settings varying according to the model. Oversized settings may manifest issues of model non-convergence. During construction, the training data should undergo multiple shuffling processes. Sole reliance on enabling shuffle solely through the configuration file may prematurely fit the model, leading to overfitting problems characterized by sensitivity to some tasks while incapacity for others. Additionally, in experimenting with the base model for instruction fine-tuning, it was observed that using instruction fine-tuning data does not enhance the base model's effectiveness. On the contrary, as fine-tuning progresses, the model's performance deteriorates gradually. Ultimately, this degradation may lead to a complete absence of output or output characterized by garbled information.

\paragraph{Problems during training}
In the aforementioned experiments, it was noted that the training data lacked multiple shuffling instances, leading to premature model fitting. To comprehensively investigate this phenomenon, a more in-depth analysis was undertaken. Specifically, secondary fine-tuning of instruction data was performed on the checkpoint of the pre-fitted model, aiming to address the issue of the model's sensitivity to certain tasks while lacking the ability to solve others. Attempts were made using remaining untrained instruction fine-tuning data, task data beyond the model's capabilities, LoRA fine-tuning, and other approaches; however, none successfully resolved the identified issues. Furthermore, to explore whether these issues stem from data problems or training methods, a series of controlled experiments was conducted. This involved retraining after complete data disruption, concluding when reaching the same number of steps, followed by a repetition of the training process. The results consistently indicated that secondary training consistently led to poorer model output.

\subsection{Model training experience}
The model training process exhibits minimal variation between the Ascend server and the NVIDIA graphics card, with both sides encountering essentially identical issues. The purpose of this report is to delve into the challenges faced during project development and share insights into the corresponding solutions. Therefore, this section predominantly addresses the data-related challenges encountered during training. Additionally, it outlines our application of QLoRA fine-tuning, utilizing position interpolation to address issues associated with processing lengthy text.
\paragraph{Data issues}
The most critical data issue encountered during training pertained to the noise within the aforementioned KnowLM-IE dataset. Despite the dataset's substantial size, nearing 30W data points, a significant portion is acquired through remote supervision, introducing considerable noise. Moreover, the KnowLM-IE dataset encompasses diverse fields, including people, geographical areas, transportation, organizations, works, organisms, events, etc. While it functions as a triple extraction dataset, distinctions exist between this dataset and our standard triple extraction data. Using this dataset directly for instructive fine-tuning poses a risk of compromising the model's ability to handle entity-relation triples post-training. To address this, a portion of the data was filtered and incorporated as newly added triplet data.Additionally, during the initial stages, researchers in our group annotated data for both the RTE task and the NER task, assigning answers to both tasks on the same data instances. Consequently, post-training, the model exhibited certain confusion issues between NER and RTE tasks, manifesting as instances where a question pertains to RTE, and the response corresponds to the NER task.

\paragraph{long text problem}
As the prevalence of large models increases, accompanied by the handling of more intricate tasks, there is a growing necessity for models to effectively process longer conversations and documents while sustaining optimal performance in original, shorter input tasks. Due to the attention mechanism's limitation in the Transformer architecture, where computational complexity exhibits a quadratic relationship with input text length, the context window of pre-trained large models remains relatively fixed, restricting its applicability to longer texts. Addressing this challenge has become a focal point in current research, with mainstream methods encompassing direct extrapolation and position interpolation. In the TechGPT-2.0 project, our model employs the RoPE\cite{su2023roformer} method, relying on relative distance for attention score computation. Direct extrapolation, due to its potential for greater confusion issues, is eschewed in favor of the position interpolation method, enhancing the model's capacity to handle long texts. The key concept involves scaling the longer position index to the original window range, ensuring the maximum relative distance between any two tokens does not exceed the original window size. This mitigates the extended window's impact on attention score calculation. Importantly, the position interpolation method introduces minimal modifications to the model structure, requiring only a few adjustments in RoPE. This design choice renders the extended model compatible with the original model's infrastructure and optimization methods, enhancing practical applicability. Additionally, to facilitate the model's adaptation to the expanded window size, we constructed a limited number of long text samples for fine-tuning the extended model. Notably, in addressing the long text problem, the Ascend server was not utilized; instead, the QLoRA method extended the TechGPT model's window to up to 12K length. Evaluation across multiple long text tasks demonstrates the extended model's reasoning capability for longer texts while maintaining performance in original, shorter text tasks.


\section{Conclusion and future work}
In this report, we present the recent TechGPT-2.0 project undertaken by our research group, encompassing two 7B-scale models for knowledge graph construction tasks and a QLoRA weight tailored for addressing long-text issues. Emphasis is placed on evaluating the performance of large models with reduced parameters in knowledge graph construction tasks, aiming to furnish the Chinese open-source community with a substantial model capable of constructing knowledge graphs while retaining the overarching capabilities of the Chat model. All models within this project are derived from LLAMA2, initially pre-trained in Chinese, and subsequently fine-tuned using a substantial set of instructions. The Chat version, demonstrating commendable performance, serves as the initial model and undergoes further refinement through instructions derived from the collected knowledge graph construction task dataset. This report serves to outline challenges encountered during the TechGPT-2.0 project's development and the resultant insights. It is intended solely as a reference for subsequent researchers; hence, extensive experimental demonstrations have not been conducted. Specifically, the report delves into nuances in Ascend Server utilization for model selection, intricacies in the data collection and sorting process, challenges faced during Ascend Server-based training, and the derived experiences. It also addresses issues encountered in model training and the technical solutions applied to address long-text challenges.

Although a small-scale verification test was conducted on TechGPT-1.0, the experiments remain insufficient for the comprehensive evaluation of the TechGPT-2.0 project. Subsequently, we plan to augment the report with additional details regarding model implementation and conduct more intricate experiments on the developed model, illustrating the experiences articulated in this report. Comprehensive experiments on the model, encompassing knowledge graph construction task results, toxicity tests, comparative experiments, and related evaluations, will be conducted to further underscore the model's performance. Moreover, the models within this project will be leveraged for additional research endeavors, including RAG\cite{lewis2020retrieval}, Agent, tool calling, and multi-modality studies.


\bibliographystyle{unsrt}  
\bibliography{references}

\end{document}